\documentclass{bmvc2k}
\usepackage{amsfonts}
\usepackage{wrapfig}
\usepackage[normalem]{ulem}
\usepackage{array}
\usepackage{graphics}

\title{TalkLoRA: Low-Rank Adaptation for Speech-Driven Animation}

\addauthor{Jack Saunders}{https://jsaunders909.github.io/}{1, 2}
\addauthor{Vinay P Namboodiri}{https://vinaypn.github.io/}{1}

\addinstitution{
 University of Bath \\
 Bath, UK
}
\addinstitution{
 Deepreel Ltd \\
 124 Goswell Rd,\\
 London, UK
}

\runninghead{Saunders et. al.}{TalkLoRA}

\begin{document}

\maketitle

\begin{figure}[!h]
    \centering
    \includegraphics[width=\textwidth]{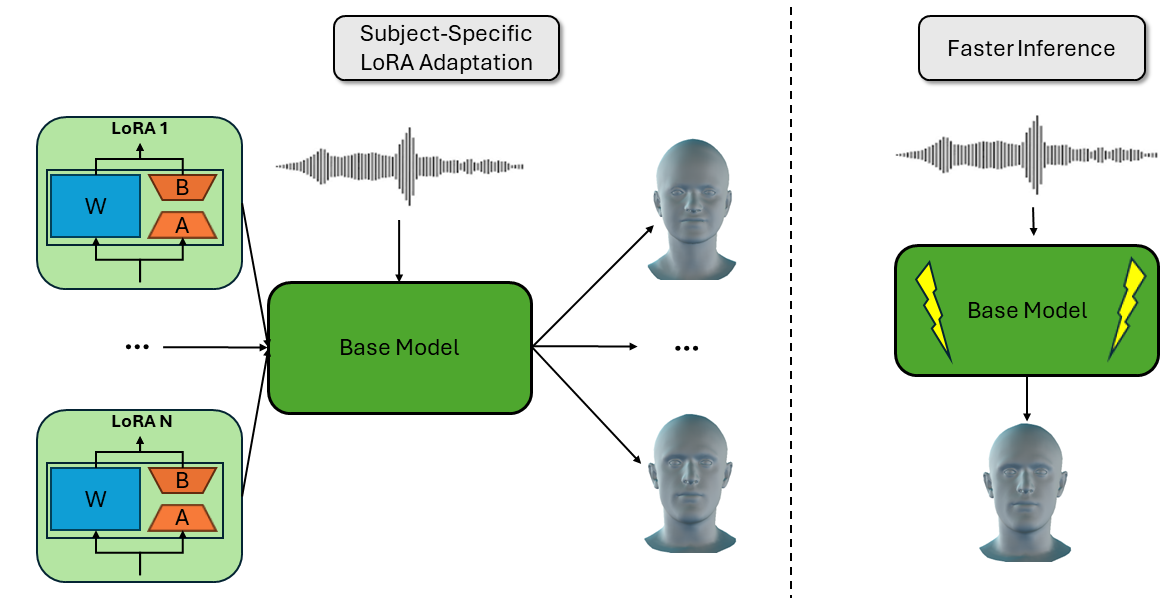}
    \caption{We present TalkLoRA, a method for improving any transformer-based speech-driven animation model. We use Low Rank Adaptation to effectively and efficiently adapt to new identities and chunking to improve inference speed, with no loss of quality.}
    \label{fig:title}
\end{figure}

\begin{abstract}
Speech-driven facial animation is important for many applications including TV, film, video games, telecommunication and AR/VR. Recently, transformers have been shown to be extremely effective for this task. However, we identify two issues with the existing transformer-based models. Firstly, they are difficult to adapt to new personalised speaking styles and secondly, they are slow to run for long sentences due to the quadratic complexity of the transformer. We propose TalkLoRA to address both of these issues. TalkLoRA uses Low-Rank Adaptation to effectively and efficiently adapt to new speaking styles, even with limited data. It does this by training an adaptor with a small number of parameters for each subject. We also utilise a chunking strategy to reduce the complexity of the underlying transformer, allowing for long sentences at inference time. TalkLoRA can be applied to any transformer-based speech-driven animation method. We perform extensive experiments to show that TalkLoRA archives state-of-the-art style adaptation and that it allows for an order-of-complexity reduction in inference times without sacrificing quality. We also investigate and provide insights into the hyperparameter selection for LoRA fine-tuning of speech-driven facial animation models.
\end{abstract}

%-------------------------------------------------------------------------
\section{Introduction}
\label{sec:intro}

3D Digital humans are very pervasive across many forms of media. TV, video games, movies, telepresence and marketing, make extensive use of them. Furthermore, they are a critical component in 2D Talking Head generation \cite{saunders2023read, Saunders2024D4E, thies2020nvp, thies2019deferred, Ma_2023_AAAI}. As social creatures, humans pay a lot of attention to each other's faces \cite{Dobs2018-sm}. This makes us very good at discerning details relating to the face. Of particular importance is the motion of the face. With even small errors in this facial animation, the end result enters into what is known as the `uncanny valley', an unsettling phenomenon that prevents acceptance of the digital human \cite{mori2012uncanny}. 

A traditional way of obtaining high-quality facial animations is for skilled artists to manually pose the face into keyframes, and to interpolate between these. This process is, however, very slow and expensive, making it feasible only for the most important facial animation. Another popular method is the use of performance capture which involves attempting to match an actor's performance to a 3D face rig (e.g. \cite{DECA, EMOCA:CVPR:2021, SMIRK:CVPR:2024}). This process is, again, costly. For high-volume facial animations, speech-driven facial animation is popular. Furthermore, certain applications require \textit{procedurally generated} facial animations that are synthesised on the fly from audio or text input. For example text-driven avatars, or conversational agents.

A recent surge of methods has excelled at producing high-quality facial animations from speech signals \cite{Imitator, FaceFormer, EMOTE, aneja2023facetalk, VOCA2019, richard2021meshtalk}. These methods, however, mostly require a significant amount of data per person. To date, only a small number of works have considered how best to adapt speech-driven animation systems to new identities \cite{Imitator}. 

Furthermore, a large number of these works are transformer-based. This means to generate a facial expression at time $t$, they look at all tokens in the range $[0, t-1]$, making them $O(N^2)$ in complexity where $N$ is the length of the animation. This is unsuitable for long animations. It is also a counterintuitive approach, conditioned on audio, why should the facial expression at any given time depend on the expression from, for example, ten seconds ago? The use of full-length context is therefore unnecessary and detrimental.

We propose \textbf{TalkLoRA} to address both these problems. TalkLoRA uses Low-Rank Adaptation to adapt existing pre-trained transformer speech-driven models. LoRA allows for more efficient and effective fine-tuning in both limited and plentiful data scenarios. Furthermore, TalkLoRA utilizes a fixed context window in the transformer, reducing the computational complexity to a constant level and allowing much longer sequences to be processed without a drop in quality. 

It is important to note that while we base TalkLoRA on two specific models, it could equally be applied to any transformer-based model, including emotional models \cite{EMOTE} or even speech-driven implicit models \cite{aneja2023facetalk, giebenhain2023nphm}. 

In summary, our contributions are:

\begin{itemize}
    \item{A methodology for effective and efficient adaptation of any transformer-based speech-driven animation model (Table \ref{tab:main_results}) that runs with an order-of-complexity reduction in inference time (Section \ref{sec:chunk}).}
    \item{A LoRA-based adaptor for fitting speech-driven animation models to new subjects, together with an analysis on the improvement over state-of-the-art (Table \ref{tab:main_results}), the effect of dataset size (Table \ref{tab:main_results}) and the impact of hyperparameter selection (Sec \ref{sec:LoRAHyperparameters}).}
    \item{The use of a limited context window in the transformer through chunking, that reduces the inference time. We also include analysis of this component, showing that it does not reduce quality, and an investigation into the required context length (Figure \ref{fig:chunking} \& \ref{fig:chunking_results}).}
\end{itemize}

\section{Related Work}

\subsection{Speech-Driven 3D Facial Animation}

Speech-driven 3D facial animation involves producing facial animations, as sequences of either blendshapes or vertices from an input audio sequence. Very early methods use rule-based approaches based on visemes (the visual counterpart to phonemes) \cite{jali16, Kalberer01, DEMARTINO2006971, Cosker04}. These methods work fairly well for high-volume digital humans, but are very simplistic and cannot faithfully produce high-quality animations. Since the advent of deep learning, many data-driven methods have been proposed with greater capabilities \cite{Audio2Face, Imitator, FaceFormer, EMOTE, VOCA2019}. Karras et. al. \cite{Audio2Face} train an auto-regressive CNN-based model to predict vertices from audio segments. This method only works on the actor that it is trained for. VOCA \cite{VOCA2019} improves upon this by adding one network shared across several identities, disambiguation is achieved through a one-hot encoding of identity, and novel identities can be added using a linear interpolation of existing ones. Faceformer \cite{FaceFormer} works similarly with one-hot encodings and achieves better results with the use of transformers. Several works apply this transformer-based, one-hot identity model to subdomains, for example, emotional speech-driven animation \cite{EMOTE} or speech-driven animation of implicit models \cite{NPHM, aneja2023facetalk}. There are two significant drawbacks to this. First, the one-hot encoding severely limits the ability of the model to adapt to new identities. Secondly, the transformer decoders use the entire audio segment for generation with the self-attention mechanism. This quickly becomes infeasible for long sentences. Imitator \cite{Imitator} looks to address the first of these problems by fine-tuning certain layers in their model. While this does help, it is both slow and sub-optimal. Our method, by comparison, is able to effectively and efficiently adapt to new identities using LoRA \cite{hu2022lora} and can handle longer sequences using a fixed-context window.

\subsection{Transfer Learning}

Transfer Learning is a very general problem in the field of machine learning. The objective is to take a large model that has been trained on one dataset and to adapt it for use on another, often smaller, dataset. Transfer learning has been most prominently investigated in the context of NLP (e.g. \cite{LLMFewShot, radford2018improving, Peft19}) and vision \cite{xin2024parameter, chen2022adaptformer, jie2022convpass}. Recent work has also investigated the topic of speech recognition \cite{Xu2021SimpleAE}, which is closely related to our work on speech-driven animation.

A naive way of achieving transfer learning to resume the training of a model using the pre-trained weights and optimising all parameters. However, this has several issues. For one, this is a slow process and it is very memory intensive. Furthermore, if the new dataset is small, training a large model in this way makes it very vulnerable to overfitting. This has given rise to the field of parameter-efficient fine-tuning (PEFT), where only a small number of parameters are updated. Some methods focus on tuning a subset of network layers, however, this restricts the level of abstraction at which the tuning operates. More recently, methods have addressed this problem by optimising more layers using lower dimensional decomposition with small neural networks \cite{Peft19}. Unfortunately, this introduces a large overhead at inference time. Therefore Low-Rank Adaptors (LoRA) were proposed \cite{hu2022lora}. LoRA works by decomposing weight matrices into two lower-rank matrices whose product is the same shape as the weight matrix. This can be added to the pre-trained weights and used with no overhead, yet still allowing for parameter-efficient finetuning.

In the context of speech-driven facial animation, the goal of transfer learning is to adapt pre-trained models to new identities. Often, there is little person-specific data available on which to perform adaptation. This means that it is essential to avoid overfitting. Imitator \cite{Imitator} achieves this by fine-tuning a style code and the final layer. This works well but is highly restrictive, effectively serving as an interpolation of existing styles with an additional linear transformation. We propose using LoRA \cite{hu2022lora} which allows us to adapt the model in a non-linear and more flexible way, leading to better results.

\section{Method}

Our objective is to improve any existing transformer-based speech-driven animation system. We therefore propose components that do not make any additional assumptions about the model. We first describe two existing state-of-the-art speech-driven animation models (Section \ref{sec:arch}) to show the differences, our model works on both. We then discuss our use of Low-Rank Adaptors to allow us to adapt existing models to new identities (Sec \ref{sec:lora}). Finally, we improve the inference speed of existing models for long sequences using a chunking strategy (Sec \ref{sec:chunk}). We show these steps in Figure \ref{fig:title}.

\subsection{Architectures}
\label{sec:arch}

The basic architecture of our model is determined by the base model we use for adaptation. For the case of our experiments this is either FaceFormer \cite{FaceFormer} or Imitator \cite{Imitator}. In either case, there are some significant similarities. Each model consists of three components, an audio encoder, a transformer decoder and an per-frame decoder.

\noindent \textbf{Audio Encoder:} For both Imitator and Faceformer the audio encoder is the same. Wav2Vec2 \cite{wav2vec220} is used due to its powerful ability as a feature extractor. Wav2Vec2 is trained for speech recognition on a large and diverse dataset. The choice of speech recognition as a task means that the output features of the model are person-agnostic, allowing for good generalisation of the speech-driven models to novel audio. The final layer is discarded and the outputs of the final hidden layer are taken as the audio features. The Wav2Vec2 model outputs features at 50Hz, these are linearly resampled to match the fps of the target animations, and a learnable linear layer converts these to the desired dimensionality.

\noindent \textbf{Tranformer Decoder} Both models also make use of a transformer decoder to consider temporal information. They each use cross-attention using the audio features from the audio encoder. The key difference is that Imitator's transformer is person-independent while FaceFormer's is not. FaceFormer encodes the vertices from the previous time step, adds a person-specific style code and then passes the result to a transformer. Whereas Imitator uses a predefined start token, which is the same for all subjects, and produces person-independent viseme tokens from the audio features. It is important to note that TalkLoRA does not require either of these choices over the other, both our adaptor and the chunking procedure can be applied to any transformer-based model.

\noindent\textbf{Motion Decoder} The final step is to produce the vertices from the transformer output. As FaceFormer includes style information in the transformer, only a single linear layer is used to project the transformer output onto the FLAME vertices. Imitator, on the other hand, introduces the person-specific style code here by adding it to the transformer output. Imitator then uses an MLP to predict vertices.

\subsection{LoRA}
\label{sec:lora}

In order to adapt a baseline model to a new subject we make use of Low-Rank Adaptors (LoRA) \cite{hu2022lora}. LoRA is a method for parameter-efficient fine-tuning originally proposed for large language models. Instead of training all the weights in a model, or some subset of the layers, LoRA adds an offset to the weight matrices of layers using rank decomposed matrices. Specifically, consider a weight matrix $W \in \mathbb{R}^{N \times M}$. LoRA models the adaption as:

\begin{equation}
    W' = W + \Delta W = W + AB^T
\end{equation}

\noindent Where $A \in \mathbb{R}^{N \times r}$ and $B \in \mathbb{R}^{M \times r}$, with $r << \min(N, M)$. The idea is that as new datasets have low intrinsic dimensionality, the required weight updates will also.

We have several considerations for which components of the network we should apply LoRA to. First, for the audio encoders, we consider it counterintuitive to apply LoRA. The audio encoder is powerful precisely because it is highly generalised. This allows it to encode audio from any person (for example from a TTS system or voice actor) into a common feature space. While adapting the audio encoder may improve the performance for audio coming from the subject, it is likely to overfit and hinder performance from other audio sources. We therefore do not consider applying LoRA to this part of the network.

For the decoders, however, we want the model to adapt to just a single identity. That is, we want the decoded motion to take on the style of the speaker. This gives us two candidates for LoRA application. The transformer decoder and/or the motion decoder. We show which is preferable in Section \ref{sec:LoRAHyperparameters}.

LoRA introduces a small set of parameters for fine-tuning. This allows for a tradeoff between the representational power of the model and regularization. That is, how likely it is to overfit. Given a dataset with a lower intrinsic dimensionality, we would expect to use a smaller value of $r$. In Section \ref{sec:LoRAHyperparameters}, we determine the optimal value of $r$ for our datasets empirically.

\subsection{Limiting Context Window for Speedup}
\label{sec:chunk}

\begin{figure}
    \centering
    \includegraphics[width=\textwidth]{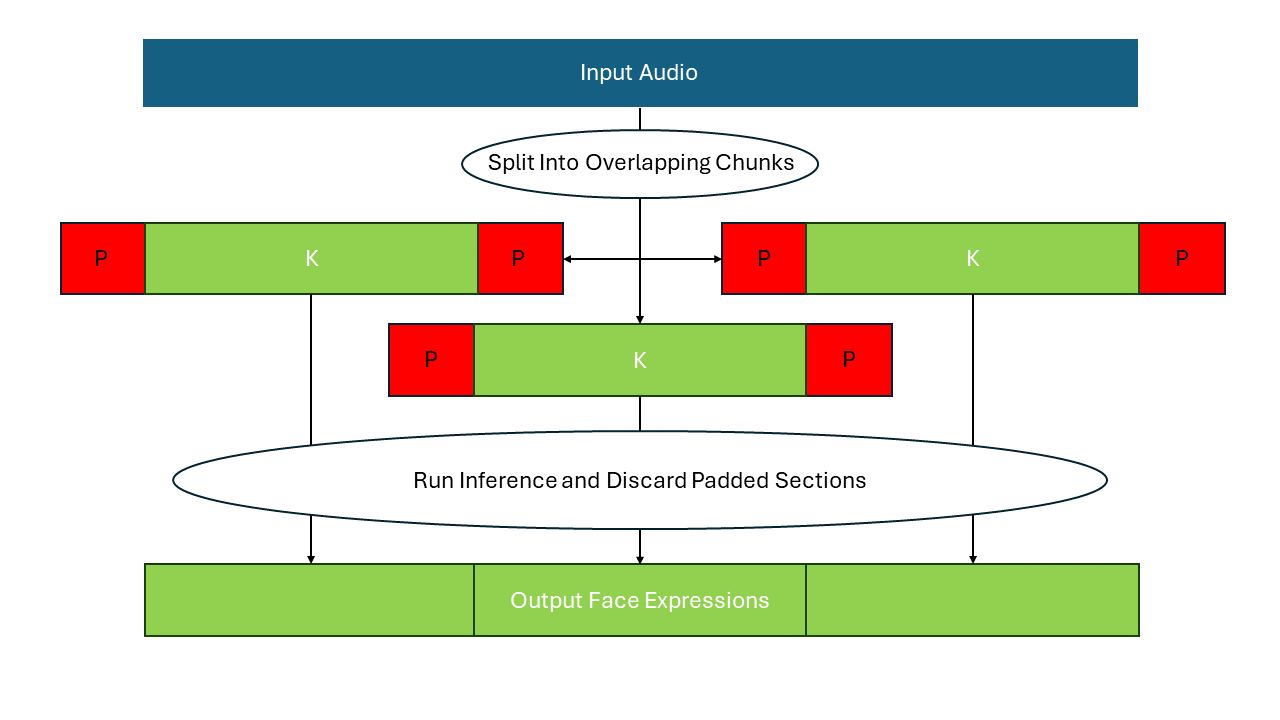}
    \caption{The chunking process is used to limit the context window of the transformers. We split incoming audio into overlapping chunks of size K+2P and process these in parallel. The padding is then removed and the results concatenated.}
    \label{fig:chunking}
\end{figure}

Faceformer \cite{FaceFormer} and Imitator \cite{Imitator}, as well as other transformer-based speech-driven animation methods \cite{EMOTE, aneja2023facetalk}, all use transformers with unlimited context lengths. To compute what the lips should look like at time $t=60s$, they will look at the entire history of the lip motions from time $t=0s$. This is a holdover from the original use of transformers in NLP, where words may hold influence over long time periods. For facial animation, however, this assumption does not make sense. The audio is used as input, meaning there is no need for the transformer to learn an internal language model, only the audio in a short context window, and the last few frames of the facial expression should matter.

We therefore look to reduce the context window of the transformers. \textbf{Note that we are not retraining the base models}. Therefore, we need to alter the architecture at inference time in such a way that it only sees a small context window without degrading performance.

To do this, we apply chunking. We split the input audio of length $T$ into overlapping padded chunks of fixed size $K + 2P$. Here $P$ is the size of the padding. The idea is that the padding ``warms up" the transformer, providing good context for inference. The predictions in this padded region are likely to be poor, so these are simply discarded. An example of this is shown in Figure \ref{fig:chunking}.

Transformers notoriously have quadratic complexity in the size of the input sequence ($\mathcal{O}(N^2)$. By using a constant and fairly small value $K < N$ the complexity is simply a linear $\mathcal{O}(NK^2) << \mathcal{O}(N^2)$ ($\frac{N}{K}$ sequences of length K). This is an order of complexity reduction. We determine the values of $K$ and $P$ required for optimal performance in Section \ref{sec:chunkhyperparameters}.

\section{Implementation Details}

For each base model, we train using the procedures set out in the respective papers. We use these base models as a baseline. For person-specific adaptation we use the loss weights outlined in Imitator \cite{Imitator} with $\lambda_{\textit{rec}}=1.0$ and $\lambda_{\textit{vel}}=10.0$. We use the AdamW optimiser with a learning rate of $0.001$. Unless otherwise specified, we use a LoRA rank of 4 and LoRA alpha value of 8. We find empirically that the model converges after just $50$ epochs, so we train our adaptor for this many.

\section{Results}

\begin{table}
\centering
\caption{Comparison of adaptation strategies, we compare our method to Imitator's described adaptation \cite{Imitator}, and FaceFormer with adaptation of the style code \cite{FaceFormer}. We show that our adaptation strategy improves both models. We also include the baseline models for reference. The best results for each dataset size are shown in \textbf{bold}, the best adaptation for each base model and dataset combination is \underline{underlined}.}
\label{tab:main_results}
\resizebox{\columnwidth}{!}{%
\begin{tabular}{c|cccc|cccc}
                                  & \multicolumn{4}{>{\centering\hspace{0pt}}m{0.369\linewidth}|}{Subject A}                          & \multicolumn{4}{>{\centering\arraybackslash\hspace{0pt}}m{0.369\linewidth}}{Subject B}             \\
                                  & $L_2^{Face}\downarrow$& $L_2^{Lip}\downarrow$& Lip-Max $\downarrow$& Time (mm:ss)$\downarrow$& $L_2^{Face}\downarrow$& $L_2^{Lip}\downarrow$& Lip-Max$\downarrow$& Time (mm:ss)$\downarrow$\\ 
\hline
Faceformer (Base)                 & 0.933                  & 0.181                  & 6.097                  & -                      & 0.710                  & 0.098                  & 3.694                  & -                       \\
Imitator (Base)                   & 0.902                  & 0.157                  & 5.609                  & -                      & 0.736                  & 0.141                  & 4.781                  & -                       \\ 
\hline
Imitator (1 Sentence)             & 0.911                  & 0.136                  & 5.021                  & 07:29& 1.065                  & 0.091                  & 3.462                  & 10:43                   \\
Imitator + Ours (1 Sentence)      & \uline{0.900}          & \textbf{\uline{0.135}} & \textbf{\uline{4.975}} & \textbf{\uline{00:30}}& \uline{1.059}          & \textbf{\uline{0.090}} & \textbf{\uline{3.440}} & \textbf{\uline{00:42}}\\
Faceformer + Style (1 Sentence)   & 0.881                  & \uline{0.153}          & 5.397                  & 03:46& 0.715                  & 0.118                  & 4.160                  & 04:27\\
Faceformer + Ours (1 Sentence)    & \textbf{\uline{0.876}} & \uline{0.153}          & \uline{5.395}          & \uline{00:40}& \textbf{\uline{0.698}} & \uline{0.114}          & \uline{4.074}          & \uline{00:50}\\ 
\hline
Imitator (3 Sentence)             & 0.904                  & 0.136                  & 4.946                  & 24:23                  & 1.114                  & \textbf{\uline{0.094}} & 3.515                  & 34:57                   \\
Imitator + Ours (3 Sentence)      & \uline{0.903}          & \textbf{\uline{0.134}} & \textbf{\uline{4.884}} & \textbf{\uline{01:55}}& \uline{1.113}          & \textbf{\uline{0.094}} & \textbf{\uline{3.157}} & \textbf{\uline{02:17}}\\
Faceformer + Style (3 Sentences)  & 0.875                  & 0.151                  & 5.366                  & 12:13                  & 0.707                  & 0.114                  & 4.025                  & 13:50                   \\
Faceformer + Ours (3 Sentences)   & \textbf{\uline{0.852}} & \uline{0.144}          & \uline{5.248}          & \uline{02:14}& \textbf{\uline{0.811}} & \uline{0.97}           & \uline{3.790}          & \uline{02:29}\\ 
\hline
Imitator (5 Sentence)             & 0.878                  & 0.131                  & 4.780                  & 50:35                  & \uline{1.070}          & 0.093                  & \textbf{\uline{3.481}} & 61:03                   \\
Imitator + Ours (5 Sentence)      & \uline{0.877}          & \textbf{\uline{0.128}} & \textbf{\uline{4.657}} & \textbf{\uline{03:19}}  & 1.103                  & \textbf{\uline{0.092}} & 3.490                  & \textbf{\uline{04:07}}   \\
Faceformer + Style (5 Sentences)  & 0.859                  & 0.149                  & 5.359                  & 20:45                  & \uline{\textbf{0.703}} & 0.110                  & 3.936                  & 23:40                   \\
Faceformer + Ours (5 Sentences)   & \uline{\textbf{0.838}} & \uline{0.139}          & \uline{5.137}          & \uline{03:49}           & 0.830                  & \uline{0.098}          & \uline{3.863}          & \uline{04:19}            \\ 
\hline
Imitator (10 Sentence)            & \uline{0.867}          & 0.129                  & 4.659                  & 93:39                  & 0.865                  & 0.086                  & \textbf{\uline{3.244}} & 107:22                  \\
Imitator + Ours (10 Sentence)     & 0.888                  & \textbf{\uline{0.126}} & \textbf{\uline{4.547}} & \textbf{\uline{06:33}}& \uline{0.783}          & \textbf{\uline{0.086}} & 3.252                  & \textbf{\uline{07:45}}\\
Faceformer + Style (10 Sentences) & 0.848                  & 0.147                  & 5.340                  & 39:29                  & \textbf{\uline{0.697}} & 0.098                  & \uline{3.726}          & 43:32                   \\
Faceformer + Ours (10 Sentences)  & \textbf{\uline{0.816}} & \uline{0.133}          & \uline{4.960}          & \uline{07:18}& 0.705                  & \uline{0.094}          & 3.731                  & \uline{07:53}\\ 
\hline
Imitator (30 Sentence)            & 0.697                  & 0.125                  & 4.470                  & 259:19                 & 0.635                  & 0.083                  & \textbf{\uline{3.058}} & 316:54                  \\
Imitator + Ours (30 Sentence)     & \textbf{\uline{0.695}} & \textbf{\uline{0.124}} & \textbf{\uline{4.367}} & \textbf{\uline{17:07}} & \textbf{\uline{0.595}} & \textbf{\uline{0.082}} & 3.130                  & \textbf{\uline{21:00}}  \\
Faceformer + Style (30 Sentences) & 0.790                  & 0.139                  & 5.08                   & 117:36                 & 0.612                  & 0.091                  & 3.631                  & 129:37                  \\
Faceformer + Ours (30 Sentences)  & \uline{0.739}          & \uline{0.125}          & \uline{4.677}          & \uline{20:43}          & \uline{0.611}          & \uline{0.90}           & \uline{3.604}          & \uline{23:23}           \\
\hline
\end{tabular}
}
\end{table}

\begin{figure}
    \centering
    \includegraphics[width=\textwidth]{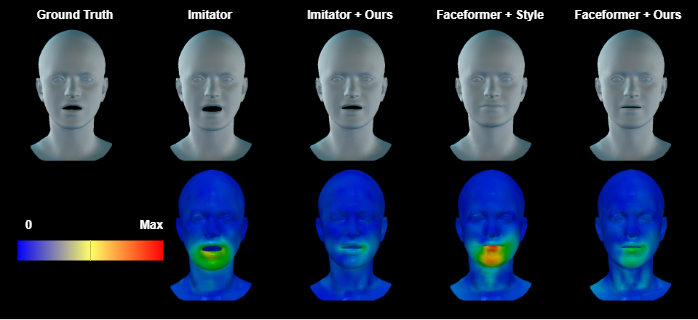}
    \caption{Qualitative results of our method showing a sentence on one of the train subjects. We compare our adaptation method on both Imitator and Faceformer and show improvements over their respective adaptation methods.}
    \label{fig:vis}
\end{figure}

\subsection{Data}

\begin{figure}[t]
    \centering
    \includegraphics[width=\textwidth]{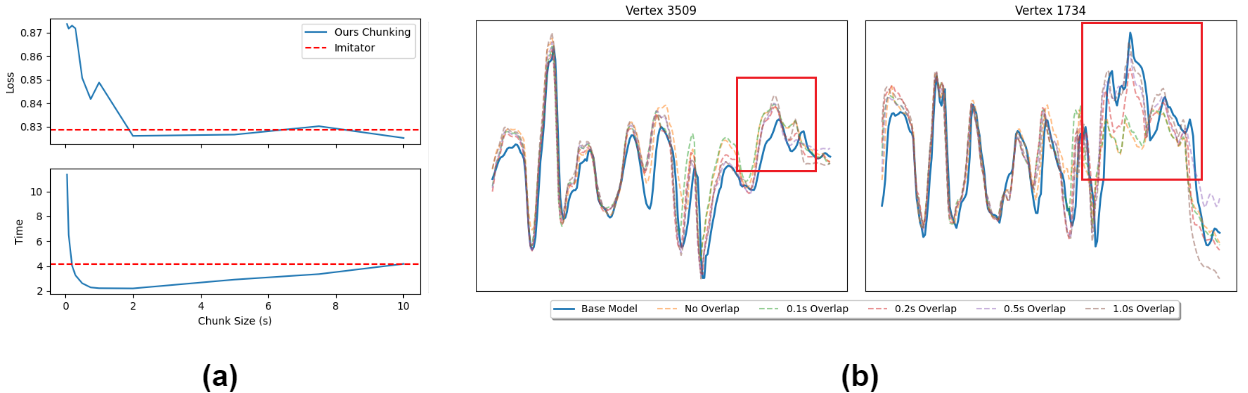}
    \caption{Graphs for determining the values of chunk size (K) and padding size (P) for chunking. \textbf{(a)} shows the effect of the size (K) of chunks compared vs the inference time and the validation loss for a validation subject. Too small a chunk takes a long time due to the padding, and also has poor quality. We find a sweet spot for time savings and quality at around 1-3 second chunks. \textbf{(b)} shows the effect of overlap size (P) in chunking. We show the y-postion of two lip vertices over time. It can be seen that a 0.2s overlap in chunking allows for outputs that are close to the un-chunked base model. }
    \label{fig:chunking_results}
\end{figure}

For all of our experiments we use VOCASET \cite{VOCA2019}. VOCASET consists of the meshes from 12 subjects each speaking 40 sentences at 60fps. VOCASET is split into 8 train subjects, 2 validation subjects and 2 test subjects. For our experiments, we train the base models on the 8 train subjects and use the 2 test subjects for our person-specific adaptions. We name these test subjects Subject A and Subject B. We further split the data for subjects A and B into train and test sets. We withhold the final 10 sentences as a test set and use various subsets of the remaining 30 for adaptation depending on the experiment.

\subsection{Comparison to State-of-the-art}

To date, only Imitatior \cite{Imitator} has attempted person-specific adaptation. We therefore compare our results primarily to this model. Specifically, we compare our method of LoRA fine-tuning to Imitator's method of fine-tuning the final layer and style code of the model for 300 epochs each. We show this for several different person-specific dataset sizes, ranging from 1 sentence (about 4 seconds) up to a maximum of 30 sentences (approximately 2 minutes). Faceformer is not designed to be adapted for new identities, however, this can be done to some degree by optimising the style code used in the model. We denote this as Faceformer + Style. Following previous work, we do this for 300 epochs. In addition to this, we also include baseline results from FaceFormer and Imitator without person-specific adaptation. To do this, we run inference using each of the 8 training styles from VOCASET and select the one with the best metrics. We show results for TalkLoRA using both Imitator and Faceformer as a base model, showing that it can be applied to any transformer-based speech-driven model.

To compare models, as is standard practice, we use the $L_{2}$ distance between vertices in the last 10 sentences of the test subjects. We separate this into a full-face metric $L_{2}^{\textit{Face}}$ using all the vertices and a lip-only metric $L_{2}^{\textit{Lip}}$ that uses only the lip vertices. Following MeshTalk \cite{richard2021meshtalk} we also use the Lip-Max metric, which is defined as the mean of the maximum $L_2$ distance of any lip vertex across all frames. In addition, we also measure the time taken for each adaptation model. Specifically, we record the time to train each adaptation using an NVIDIA L4 GPU. 

The results are shown in Table \ref{tab:main_results}. Our method of adaptation achieves state-of-the-art results in the vast majority of training configurations and metrics, while being significantly faster to train. It consistently improves both Imitator and Faceformer, suggesting it can be applied to any transformer-based speech-driven animation model. The improvement is also seen visually in Figure \ref{fig:vis}. 

\subsection{LoRA Parameter Selection}
\label{sec:LoRAHyperparameters}

\begin{wrapfigure}{r}{0.5\textwidth}
\centering
\includegraphics[width=0.48\textwidth]{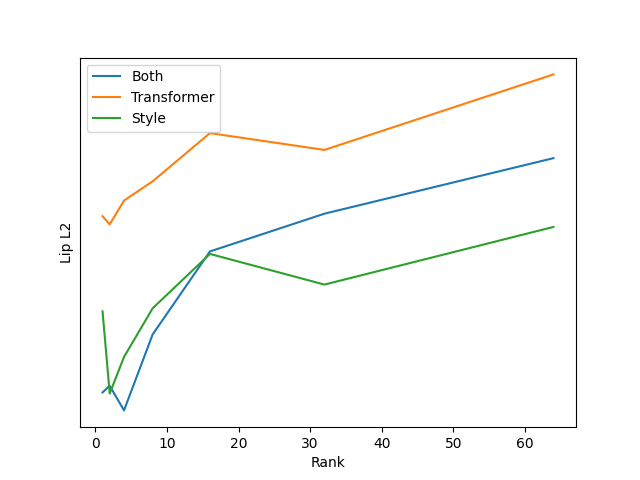}
\caption{The effect of rank on lip $L_2$ loss across random training subsets. $\approx4$ yields the best results.}
\label{fig:rank}
\end{wrapfigure}

LoRA \cite{hu2022lora} may be applied to any combination of layers in the network. Each base model is split into three components (see section \ref{sec:arch}). We find that application to the audio encoder is a bad idea, as this is specifically designed to be person-agnostic to allow for audio from any person to be used. We, therefore, only consider applying LoRA to the transformer decoder and the motion decoder. There is a single parameter that has significant influence over LoRA models, this is the rank $r$ of the decomposed matrices.

To determine the best value of rank $r$ we design a short experiment. We use the following procedure: For a random test subject, we randomly select an integer value between $1$ and $30$ representing the number of sequences we will use for fine-tuning. We then randomly select this many sequences from the training set of the given subject. We then use set $r$ to each of the values $\{1, 2, 4, 8, 16, 32 \}$ and compute the lip $L_2$ loss. We run this random sampling approach 30 times and take the average for each value of $r$. The results are shown in Figure \ref{fig:rank}. It can be seen that the optimal value for $r$ is around $4$ so this is the value we choose. Lower than this does not properly exploit the available data, while much greater means the model overfits. This suggests that the person-specific data in VOCASET has a low intrinsic dimensionality.

\subsection{Effects of Chunking}
\label{sec:chunkhyperparameters}

We also design an experiment to test the efficacy of our chunking method on long audio sequences. To do this, while still using ground truth data for computing metrics, we create artificial long sentences. This is done simply by concatenating the ten test sentences from each VOCASET subject with one second of silence in between. When calculating metrics, we mask out these silent regions. We experiment with various values of chunk size $K$ and padding $P$. We use the pre-trained base Imitator for this experiment. 

We calculate the $L_2$ loss across the long sequences for various values of $K$ and record the run time. The results of the chunking experiments are shown in Figure \ref{fig:chunking_results} (a). It can be seen that small chunk sizes cause much higher losses. This is because the transformer never gets adequate context. This effect diminishes at around 2-second long chunks. It can be seen that the run-time is optimal at 0.5-2 seconds. We therefore use a chunk size of 2 seconds for our model. For $P$ we find that the effect on the loss is much less noticeable. However, errors occur around the cut points when $P$ is too small. This can be seen in Figure \ref{fig:chunking_results} (b) in the red boxes. Small amounts of padding cause the first few frames to not have enough context leading to differences between the chunked model and the base model. We find that 0.2s of padding is enough to alleviate this without a significant increase in inference time. 

\section{Limitations and Future Work}

While our work can adapt effectively to new identities and run with faster inference, there are still some drawbacks. First, even with properly calibrated values of $K$ and $P$, chunking will still reduce the quality of models trained with long context. This means for short sentences it is often not worth chunking. In future, it may be worth investigating training models with this chunking. We hypothesise this may help prevent the model from overfitting to spurious temporal correlations in speech segments in the data. Another interesting line of research would be to include learnable weights that perform the fusing of the chunks.

While we were able to find a good set of parameters to train our LoRA models, we have only considered fixed parameters for all dataset sizes. It is likely that given more data, one may want to increase the LoRA rank as the risk of overfitting is reduced and intrinsic dimensionality increased. Exploring questions like this, as well as considering other methods of adaptation (e.g. controlnet \cite{zhang2023adding} or BOFT \cite{liu2024boft}) is left as future work.

\section{Conclusion}

We have presented our work, \textbf{TalkLoRA} for adapting transformer-based, speech-driven animation models to new identities using LoRA. Our method is capable of adapting to new speakers better than existing state-of-the-art models, as we are better able to avoid overfitting to the low intrinsic dimensionality of 3D talking head datasets. TalkLoRA also improves inference speed through our chunking strategy. Our method is applicable to any speech-driven model, provided it uses transformers, making it easy for general adoption. Our method has applications in video games and film/TV using digital characters, as well as in photorealistic, 2D, audio-driven talking head animation.

\section{Acknowledgments}

We want to thank the team at DeepReel for providing computing resources and excellent discussions on this line of research. This work was supported in part by the UKRI grant EP/S023437/1.

\bibliography{egbib}
\end{document}